\documentclass{article}

\usepackage{PRIMEarxiv}

\usepackage[utf8]{inputenc} 
\usepackage[T1]{fontenc}    
\usepackage{hyperref}       
\usepackage{url}            
\usepackage{booktabs}       
\usepackage{amsfonts}       
\usepackage{nicefrac}       
\usepackage{microtype}      
\usepackage{lipsum}
\usepackage{fancyhdr}       
\usepackage{graphicx}       
\graphicspath{{media/}}     

\title{Anime Popularity Prediction Before Huge Investments: A Multimodal Approach Using Deep Learning}

\author{
  Jesús Armenta-Segura, Grigori Sidorov \\
  Instituto Politécnico Nacional (IPN),\\ Centro de Investigación en Computación (CIC),\\ Mexico City, Mexico \\
  \texttt{\{jarmentas2022, sidorov\}@cic.ipn.mx} \\
  \\
}

\begin{document}
\maketitle

\begin{abstract}
In the japanese anime industry, predicting whether an upcoming product will be popular is crucial. This paper presents a dataset and methods on predicting anime popularity using a multimodal text-image dataset constructed exclusively from freely available internet sources. The dataset was built following rigorous standards based on real-life investment experiences. A deep neural network architecture leveraging GPT-2 and ResNet-50 to embed the data was employed to investigate the correlation between the multimodal text-image input and a popularity score, discovering relevant strengths and weaknesses in the dataset. To measure the accuracy of the model, mean squared error (MSE) was used, obtaining a best result of $0.011$ when considering all inputs and the full version of the deep neural network, compared to the benchmark MSE $0.412$ obtained with traditional TF-IDF and PILtotensor vectorizations. This is the first proposal to address such task with multimodal datasets, revealing the substantial benefit of incorporating image information, even when a relatively small model (ResNet-50) was used to embed them.
\end{abstract}


\section*{Introduction}

One of the most crucial aspects of the japanese animation industry, or \textit{anime} industry, is the release of successful and profitable products. In order to achieve this, several techniques can be employed such as cultivating a \textit{fan base}, where companies foments the \textit{popularity of the product} among a certain demography through marketing campaigns. This practice ensures a basement of devoted customers, significantly reducing the sales risk. As a consequence, the development of successful popularity prediction systems can help investors on making better decisions and can helps animation houses to avoid financial catastrophes while creating relevant, profitable and successful franchises.

When addressing this task, it comes to light that there are no straightforward methods to predict the popularity of an incoming anime. For instance, the \textit{Toei Animation}'s 1994 film \textit{Dragon Ball Z: Bio-Broly} lost more than $\yen 9$ billion \cite{dbZ-wiki,dbZ-Budget}, despite being part of the globally acclaimed \textit{Dragon Ball Z} franchise and featuring \textit{Broly}, one of its most popular antagonists. Conversely, \textit{MAPPA Studios}' 2018 anime \textit{Kimetsu No Yaiba}, based on a relatively unknown manga not among the top 50 best-selling mangas in the Oricon 2018 ranking \cite{9}, achieved unprecedented success with its 2020 film \textit{Kimetsu no Yaiba: Mugen Train}, setting a worldwide Guinness Record at the box office \cite{4}, even amidst the challenges posed by the Covid-19 pandemic. These two examples evidences that the understanding about the phenomenom is not yet enough to design deterministic approaches for efficient and accurate popularity predictors.

\begin{figure}
    \centering
    a) \includegraphics[scale=0.35]{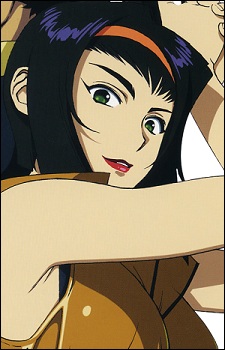} b) \includegraphics[scale=0.5]{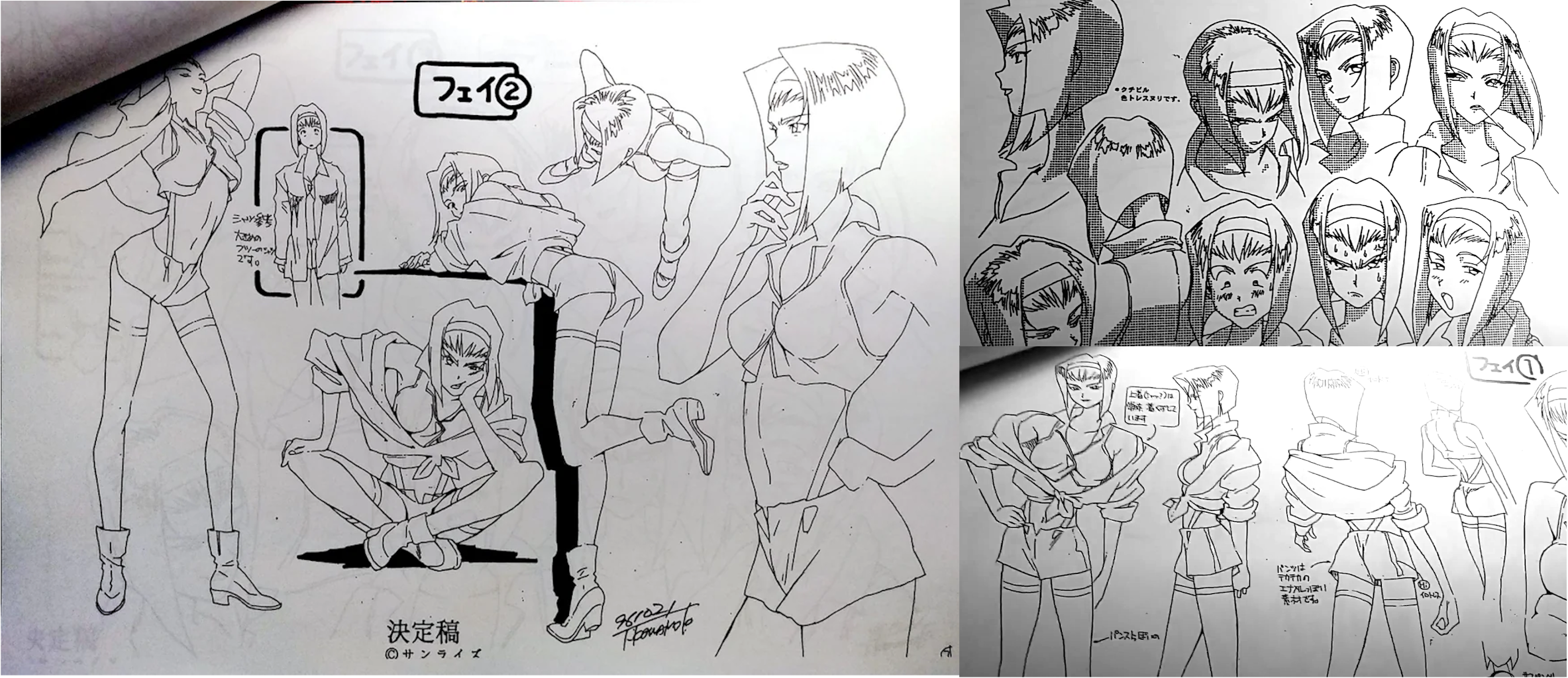}
    \caption{Visual representations of the anime character \textit{Faye Valentine}, from \textit{Cowboy Bebop}, during early stages of development. a) Her portrait in MyAnimeList. b) Her character sketch designs \cite{3} with several poses, angles and facial emotions. More than this level of detail is required for further references for animators \cite{24}, although are not assessible during early stages of development.
    }
    \label{fig:sketches}
\end{figure}

Another relevant constraint to be considered is the limited set of accessible features to design systems capable to predict popularity before making huge monetary investments. To begin with, animation has an expensive nature: even a \textit{small project} such as the pilot of the indie animation \textit{HEATHENS} is projected to cost $\$39,456$ USD (or $\$60,000$ AU) according to their crowdfunding webpage \cite{5}. Consequently, investors sometimes must make decisions even before a script is written, as is common in other entertainment industries like Hollywood, where decisions are based on a concise four-line description of the future movie plot \cite{Field2005}. In the case of animation, these four-liners might be accompanied by brief sketches of the possible main characters along with a short description of them \cite{24}. As a consequence, any system developed to predict popularity before huge investments must rely only in these limited set of features. In Figure \ref{fig:sketches} an example of such sketches is depicted.

In order to address all these quirks and constraints around the problem, this work proposes a statistical analysis fundamented on artificial intelligence (AI) methods, concretely with machine and deep learning models. In recent times, AI have become highly relevant in tackling problems of discover how a complex set of variables are related with a given dependent variable \cite{armenta2023ometeotl,maharjan-etal-2018-letting,wang2019snipets,sharma2019analysis,theodose2023kangaiset,wang2019success,wang2020fiction}. With respect of the popularity prediction task, one of the closest related AI problems is the development of recommender systems, which consists on methods that suggests the best possible products to potential users or customers \cite{Ricci2022}. When such systems are profile-based, they can determine the probability of a product being a suitable option for a consumer given their demographics or, in other words, how successful the product might be for that particular user/demographic. Although tackling the popularity prediction task through (adapted) recommender systems may be tempting, they focus on an large set of demographics which lead to a significant sparsity on their datasets \cite{Yu2023SelfSupervised}. Moreover, these datasets includes features only accessible once the product is released, such as \textit{Producers, Duration}, or \textit{Casting}. Such features have shown a strong correlation \cite{wang2019success,AIREN20231176}, but are only available in the context of a recommender system for a streaming service, leading to the need of a slightly different approach from the beginning to the end of the problem statement.

As a consequence, the first challenges for the popularity prediction task is the design of a suitable dataset. Across the internet, several anime databases containing equivalent information are available: plot synopses as free substitutes for the four-liners and in-anime snapshots with \textit{fan-made} descriptions as free substitutes for the main character sketches. The most relevant of these databases are \textit{AnimeNewsNetwork’s encyclopedia} \cite{11}, which has more than $28,000$ entries but detailed user ratings statistics for only a small subset of them; \textit{AniList} \cite{10}, with more than $15,000$ samples and detailed statistics about user ratings but strict policies against web scraping; and \textit{MyAnimeList} \cite{MAL-myanimelist}, the oldest and most active platform (according to \textit{www.similarweb.com}), encompassing more than $25,000$ entries with detailed information about user ratings for a considerable subset of them. Moreover, \textit{MyAnimeList} (MAL) does not have explicit policies against web crawling, at least until January 2024. Hence, a prudent approach to gathering freely available data must be MAL.

Once the dataset is obtained, the next challenge consists in the selection of a suitable AI model for the system. In previous experiences, there have been several proposals with other related entertainment industries, such as movies \cite{kim-etal-2019-prediction} or books \cite{maharjan-etal-2018-letting} and even in the animation industries itself \cite{armenta2023anime}. All these works have in common the employment of relatively small models with a low-to-medium complexity, but also to consider the task as a classification problem, with artificially crafted dependent variables in order to simplify the problem. From these experiences it is possible to distill the need of large and complex models as showed in \cite{armenta2023anime}, where the authors employed traditional classifiers over anime plot summaries and demonstrated the huge underlying complexity hidden on this kind of data, or as implicitly showed in \cite{kim-etal-2019-prediction}, where the authors employed a two-branched neural network with ELMO and BiLSTM or CNN to predict \textit{binary movie success} through plot summaries, but then enhanced their results by leveraging BERT-based models in \cite{MoviesBERTKimLeeCheong}.

In alignment with this distilled knowledge, this work opted for a deep neural network approach, leveraging GPT-2 \cite{radford2019language} for text analysis and ResNET-50 \cite{He2015DeepRL} for image processing. Additionally, in order to obtain the most realistic and natural results possible, the artificially crafted classifications, useful for baselines purpuses but not too fitted in reality, are set aside in favor of a regression perspective. Furthermore, this work also presents a benchmark for this model utilizing traditional vectorizations, such as TF-IDF for texts and PILtotensor for images.

In the remaining sections of the paper, all described procedures are examined in detail. The obtention of the corpus is explained, alongside the experimental setup of a three-input deep neural network used to benchmark it. To evaluate the model's performance, mean squared error (MSE) is employed to measure the error ratio of predictions. Additionally, Pearson, Spearman, and Kendall's Tau correlation coefficients are utilized to study the impact of the processed features on the score. The best results were obtained when considering all inputs, yielding an MSE of $0.011$, a Spearman correlation coefficient of $0.431$, a Pearson correlation coefficient of $0.436$, and a Kendall's Tau correlation coefficient of $0.297$.

\section*{Background}\label{Sec:Background}

\subsection*{The Anime Corpus}\label{Sec:Dataset}

For each anime in the MAL database, a python script scrapped its title, plot summary, \textit{weighted average score} and all of its main character names, descriptions and portraits as shown in Figure \ref{fig:sketches}a). All samples without this information were discarded. The scrapping process started at December 28, 2023, at 0:03 UTC, and finished at January 3, 2024, at 14:31 UTC. The script was implemented using the \textit{BeautifulSoup4} python library \cite{30}. The final result was $11,873$ animes with $21,329$ main characters.

\begin{figure}[h!]
    \centering
    \includegraphics[scale=0.5]{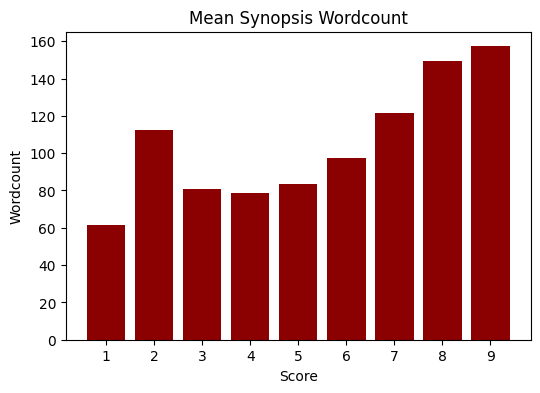}
    \includegraphics[scale=0.5]{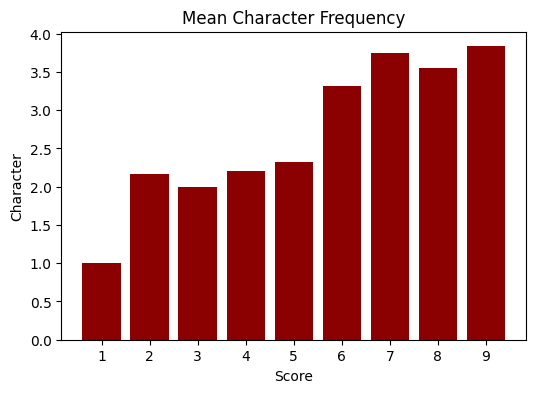}
    \caption{Mean Synopsis Wordcount left) and Mean Charcter Frequency (right) across the dataset. In both Figures the X-axis represents the floor score. Y-axis is the mean words on synopsis per score and mean amount of main characters per score.}
    \label{fig:stats_full}
\end{figure}

Once obtained the data, a clean process was performed. First, all characters with no useful descriptions, such as the text $No\ description\ available$, were removed, as well as characters with no portrait. Then, all animes with no score, synopsis, title or associated main characters were also removed, including samples with plot summaries less than $20$ words. This process reduced the scraped data to $7,784$ animes and $14,682$ characters. The characteristics of the final dataset and its statistics are presented in Table \ref{tab:stats_syn} and Figure \ref{fig:stats_full} with respect of their synopsis and in Table \ref{tab:stats_char} and Figure \ref{fig:stats_full} with respect of their main characters.

\begin{table}[h!]
    \centering
    \caption{Statistics for the corpus. Wordcount refers to synopsis.}
    \label{tab:stats_syn}
    \begin{tabular}{llllll}
    \hline
      \textbf{Score} & \textbf{Samples} & \textbf{Max. Words} & \textbf{Min. Words} & \textbf{Avg. Words}\\
    \hline
    $1-2$   & $2$     & $68$       & $55$       & $61.50$ \\
    $2-3$   & $6$     & $228$      & $25$       & $112.50$\\
    $3-4$   & $11$    & $166$      & $24$       & $80.72$ \\
    $4-5$   & $80$    & $244$      & $25$       & $78.88$ \\
    $5-6$   & $925$   & $329$      & $24$       & $83.42$ \\
    $6-7$   & $3,131$ & $397$      & $24$       & $97.35$ \\
    $7-8$   & $3,021$ & $581$      & $24$       & $121.30$\\
    $8-9$   & $596$   & $340$      & $29$       & $149.68$\\
    $9-10$  & $12$    & $189$      & $133$      & $157.25$\\
    \hline
    \textbf{Total} & $7,784$ & $581$ & $24$ & $104.73$\\
    \hline
    \end{tabular}
\end{table}

\begin{table}[h!]
    \centering
    \caption{Statistics for the characters.}
    \label{tab:stats_char}
    \begin{tabular}{llll}
    \hline
    \textbf{Total Characters} & \textbf{Max. Words} & \textbf{Min. Words} & \textbf{Avg. Words} \\
    \hline
    $14,682$ & $3,551$ & $4$ & $121.34$ \\
    \hline
    \end{tabular}
\end{table}

\subsubsection*{The MAL Weighted Average Score as Golden Label}

The most notable MAL statistic for popularity measuring is the weighted average score, calculated in terms of all user ratings of each anime. This score \cite{MAL-weighted} works straightforward: a user of the database can ranks an anime in a $0$-out-of-$10$ scale, based on their particular opinion about it. To avoid bots, the system requires the user to watch at least a fifth part of the anime before score it, which can be done by manually marking the episodes as already seen. From this, it is possible to propose a naive measure for liking as the mean of all rankings:
\begin{equation}
    S = \frac{\textrm{Sum of all users scores of the anime}}{v}.
    \label{naivescore}
\end{equation}
Where $v$ is the total number of voters. However, this measure can be biased since does not consider the statistical relevance of the population who scored it. An example of a biased score can be a hypothetical incoming anime who receive all of its first ratings from its producers, which might be interested to give generous scores, regardless the true scope of their product. To tackle this bias, MAL weighted $S$ in terms of how many people has watched and scored the anime. They defined a statistical bound $m = 50$ and they defined the weight of $S$ as:
\begin{equation}
    s = \left(\frac{v}{v+m}\right).
    \label{WeightS}
\end{equation}
Hence, if more people scores the anime, $s$ tends to $1$ and the relevance of $S$ increases. When $v=1$, the weight reaches its minimum nonzero value $1/(m+1)$, who is also the average statistical importance of a single rating.

MAL also weights the general importance of their whole community, by calculating twice per day the follow default score, based on all rankings across the database.
\begin{equation}
    C = \frac{\textrm{Sum of all valid scores in the database}}{\textrm{Total ammount of valid scores in the database}}.
    \label{Defaultscore}
\end{equation}
The value of $C$ when the data scrapping finished (Jan 3, 2024) was $6.605$. This default score represents a very coarse qualification for an incoming anime, given that nobody watched it or its fandom is statistically insignificant. Its weight is defined as follows:
\begin{equation}
    c = \left(\frac{m}{v+m}\right).
    \label{Weightdefault}
\end{equation}
When more people scores the anime, $C$ lose relevance in the score. In other words, when $s$ the weight of the scores given by the users grows, $c$ tends to $0$.

Finally, the weighted averaged score of an anime is defined as follows:
\begin{equation}
    W = \left(\frac{ v }{v+m}\right) S + \left(\frac{m}{v+m}\right) C.
    \label{Weighted}
\end{equation}
This shows the high quality of MAL metrics for popularity measuring. By considering that MAL gather members all around the world and is the most visited anime portal in all internet, any output generated by a method trained with this dataset should be interpreted as \textit{popularity across (a huge part of) the internet}. However, since this score does not consider the demographic statistics of the voters, it may be not a suitable measure for a profile-based recommender system.

\subsection*{The Deep Neural Network}

As stated in the introduction, three-input deep neural network was designed to solve the regression task of predicting the MAL Weighted Average Score given the synopsis, main character portrait and description. In Table \ref{fig:merged} an example of the full input of an anime is depicted. For each input, a sequential set of layers was employed:

\begin{table}[h!t!]
    \centering
    \begin{tabular}{ll}
    \hline
    \multicolumn{2}{c}{\textbf{Synopsis (with wordcount)}} \\
    \hline
    \multicolumn{2}{l}{Robin Sena is a powerful craft user drafted into the STNJ —a group of specialized hunters that}\\
    \multicolumn{2}{l}{fight deadly beings known as Witches. Though her fire power is great, she's got a lot to learn}\\
    \multicolumn{2}{l}{about her powers and working with her cool and aloof partner, Amon. But the truth about the}\\
    \multicolumn{2}{l}{Witches and herself will leave Robin on an entirely new path that she never expected! (66 words)} \\
    
    \multicolumn{2}{l}{}\\
    
    \hline
    \multicolumn{2}{c}{\textbf{Main Characters}} \\
    \hline
    \textbf{Name (with MAL ID)} & \textbf{Description (with wordcount)} \\
    \hline
    \textit{Robin Sena} (299)     & Robin Sena is a soft-spoken 15-year-old girl with $\dots$ (144 words)\\
    \textit{Amon} (300)           & Amon is a Hunter and is also Robin's partner. (412 words)\\
    \textit{Michael Lee} (301)    & Michael is a hacker and the technical support $\dots$ (133 words)\\
    \textit{Haruto Sakaki} (302)  & Haruto Sakaki is an 18-year-old Hunter working with the $\dots$ (167 words)\\
    \textit{Miho Karasuma} (303)  & The second in command. Miho is a 19-year-old hunter $\dots$ (73 words)\\
    \textit{Yurika Doujima} (304) & Doujima is portrayed as as carefree, lazy, vain, and $\dots$ (205 words)\\
    \multicolumn{2}{r}{\textit{\textbf{Total wordcount:} 1,134}}\\
    \hline
    \multicolumn{2}{c}{\textbf{Concatenated Portraits}}\\
    \end{tabular}

    \includegraphics[scale=1.2]{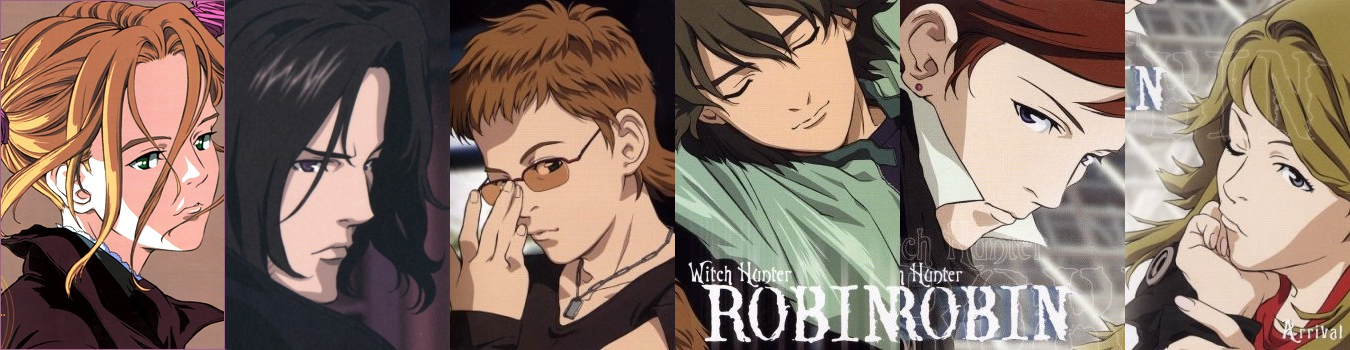}
    
    \caption{Example of the input for the anime \textit{Which Hunter Robin} with ID $7$ and Score $7.25$. }
    \label{fig:merged}
\end{table}

\begin{itemize}
    \item \textbf{Synopsis:} The GPT-2 pretrained model. \textit{Output shape: $768$.}
    \item \textbf{Main Character descriptions}: The GPT-2 pretrained model. \textit{Output shape: $768$.}
    \item \textbf{Main Character portraits}: The ResNET-50 pretrained model, flattened at the end. \textit{Output shape: $7\times7$.}
\end{itemize}

\begin{figure}[h!t!]
    \centering
    \includegraphics[scale=0.6]{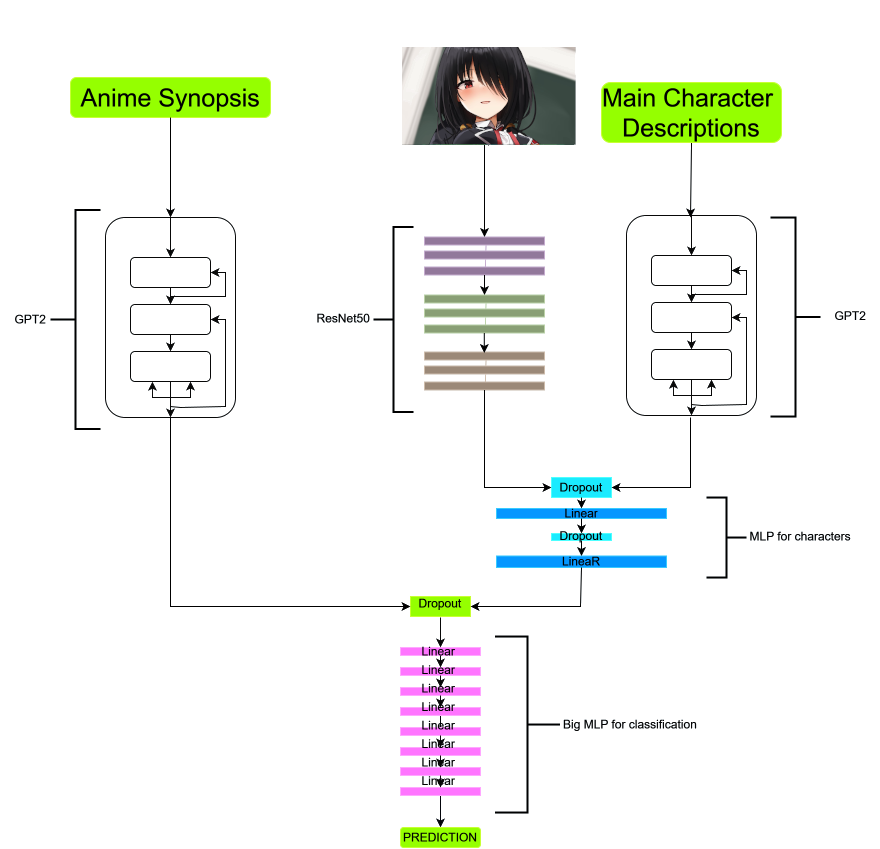}
    \caption{Full Three-input Deep Neural Network}
    \label{fig:deepNN}
\end{figure}

Once embedded with their corresponding sequential layers, the Main Character inputs were concatenated and passed to a Multilayer Perceptron (MLP) whose specifications are depicted in Table \ref{tab:smol_MLP} and Figure \ref{fig:deepNN}. The design of this MLP aims to concatenate both Main Character inputs and to process them into a unified $768$-dimentional embedding.

\begin{table}[h!]
    \centering
    \begin{tabular}{llllll}
    \hline
         \textbf{Layer Name} & \textbf{Type} & \textbf{Input Shape} & \textbf{Output Shape} & \textbf{Act. Funct.} & \textbf{Connects with} \\
    \hline

         First & Dropout $(0.1)$ & $768+49$ & $768+49$ & -- & Second \\
         Second & Linear & $768+49$ & $768$ & TanH & Third \\

         Third & Dropout $(0.1)$ & $768$ & $768$ & -- & Fourth \\
         Fourth & Linear & $768$ & $768$ & TanH & (\textit{output layer}) \\
    \hline
         
    \end{tabular}
    \caption{Specifications of the MLP for Main Characters embeddings.}
    \label{tab:smol_MLP}
\end{table}

Finally, the main character output is concatenated with the synopsis GPT-2 embeddings and passed through a larger MLP (specifications in Table \ref{tab:big_MLP}), designed to gradually reduce the dimention to convert it into a singleton suitable for regression.

\subsection*{Benchmark methods}
In order to evaluate whether the results from the deep neural network are good, the dataset is also benchmarked with traditional machine learning methods, such as in \cite{armenta2023anime}. With this purpose, all texts were vectorized through the Term Frequency-Inverse Document Frequency (TF-IDF) measure with the Scikit-Learn implementation \cite{scikit-learn}, while images were tensorized through the \textit{PILtotensor} method from the Python Imaging Library \cite{10}. The result outputs were truncated to size $750$ and then concatenated into a $2250$-dimentional tensor which was then feeded to a simple MLP described in Table \ref{tab:trad_MLP}.

\begin{table}[h!]
    \centering
    \begin{tabular}{llllll}
    \hline
         \textbf{Layer Name} & \textbf{Type} & \textbf{Input Shape} & \textbf{Output Shape} & \textbf{Act. Funct.} & \textbf{Connects with} \\
    \hline

         First & Linear & $750+750+750$ & $1000$ & TanH & Second \\
         Second & Linear & $1000$ & $500$ & TanH & Third \\
         Third & Linear & $500$ & $250$ & TanH & Fourth \\
         Fourth & Linear & $250$ & $100$ & TanH & LAST \\
         LAST & Linear & $100$ & $1$ & SoftMax & \textit{(logits)} \\
    \hline
         
    \end{tabular}
    \caption{Specifications of the MLP for the traditional methods.}
    \label{tab:trad_MLP}
\end{table}

\section*{Experiments}\label{Sec:Experiments}

In order to perform experiments, the dataset was train-test splitted in a $81:100$ proportion. The reason behind that specific ratio is that several animes can share main characters. For instance, in Table \ref{tab:Luffy}, seven out of the fifty four animes in which the main character \textit{Monkey D. Luffy} (ID 40) appears are depicted, along with the frequencies of four of his \textit{nakamas} (companions): \textit{Roronoa Zoro} (ID 62), \textit{Nami} (ID 723), \textit{Usopp} (ID 724) and \textit{Vinsmoke Sanji} (ID 305). As a consequence, it is important to ensure that all animes with shared main character belongs to the same set so all test samples will corresponds to totally unseen data.

\begin{table}[h!]
    \centering
    \begin{tabular}{llllll}
    \hline
         \textbf{Layer Name} & \textbf{Type} & \textbf{Input Shape} & \textbf{Output Shape} & \textbf{Act. Funct.} & \textbf{Connects with} \\
    \hline

         First & Dropout $(0.1)$ & $768+768$ & $768+768$ & -- & Second \\
         Second & Linear & $768+768$ & $768$ & TanH & Third \\
         Third & Linear & $768$ & $384$ & TanH & Fourth \\
         Fourth & Linear & $384$ & $192$ & TanH & Fifth \\
         Fifth & Linear & $192$ & $96$ & TanH & Sixth \\

         Sixth & Linear & $96$ & $48$ & ReLU & Seventh \\
         Seventh & Linear & $48$ & $24$ & ReLU & Eighth \\
         Eighth & Linear & $24$ & $12$ & ReLU & Ninth \\
         Ninth & Linear & $12$ & $6$ & ReLU & Tenth \\
         Tenth & Linear & $6$ & $3$ & ReLU & LAST \\
         LAST & Linear & $3$ & $1$ & SoftMax & \textit{(logits)} \\
    \hline
         
    \end{tabular}
    \caption{Specifications of the MLP for classification.}
    \label{tab:big_MLP}
\end{table}

\begin{table}[h!]
    \centering
    \begin{tabular}{llllll}
    \hline
         \textbf{Anime (with ID)} & \textbf{Luffy} & \textbf{Zoro} & \textbf{Nami} & \textbf{Usopp} & \textbf{Sanji} \\
    \hline
    One Piece (21)                          & X & X & X & X & X \\
    One Piece Film: Red (50410)             & X & X & X & X & X \\
    One Piece Movie 14: Stampede (38,234)   & X & X & X & X & X \\
    One Piece: Taose! Kaizoku Ganzack (466) & X & X & X & - & - \\
    One Piece 3D2Y: ... (25,161)            & X & - & - & - & - \\
    One Piece: Romance Dawn Story (5,252)   & X & - & - & - & - \\
    One Piece: Cry Heart (22,661)           & X & - & - & - & - \\
    \hline
    \end{tabular}
    \caption{As an example of several main characters who appears across several animes (as main characters), this table shows seven animes (out of 54) in which at least one crewmate of \textit{Monkey D. Luffy} (joined during the \textit{East Blue saga}) appears.}
    \label{tab:Luffy}
\end{table}

\subsection*{Train-Test splitting}

In order to make the custom split, all animes were grouped into clusters with respect of their shared main characters: if two animes shared a character, they were collocated in the same cluster. It is worth to note that, for each cluster, it is possible to find two animes with no shared characters, but a chain of animes who shares characters and who connects them. For that reason, the next recursive algorithm was employed to generate the clusters, obtaining $4,089$ (out of $7,784$ samples): 

\begin{itemize}
    \item For each anime, get all the other animes who shares a main character. Assign to all of them a number (cluster name).
    \item Make a second pass across all the dataset. This time, if two animes shares a cluster name, assign a new cluster name to them.
    \item Repeat this algorithm until all animes have associated a single cluster name.
\end{itemize}

\begin{figure}[h!]
    \centering
    \includegraphics[scale=0.5]{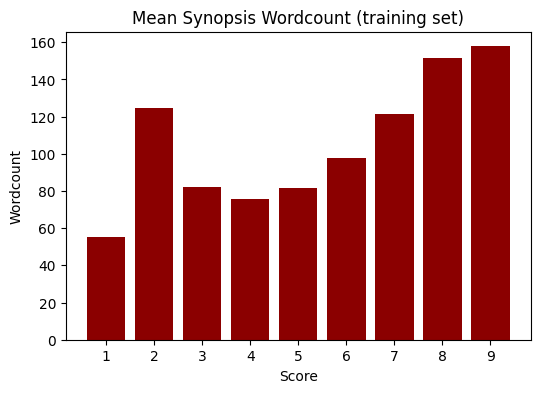}
    \includegraphics[scale=0.5]{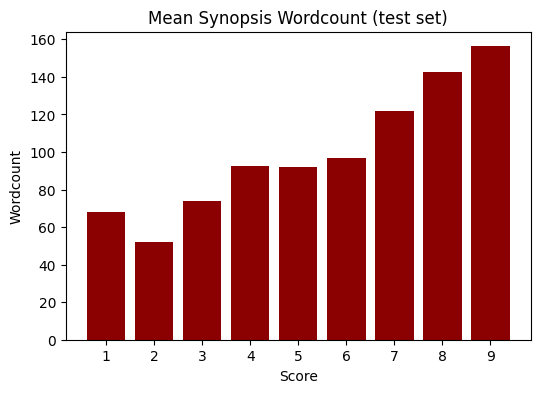}
    \caption{Mean Synopsis Wordcount in the training (left) and test (right) set. X-axis represents the floor score. Y-axis is the mean words on synopsis per score.}
    \label{fig:traintest_syn}
\end{figure}

\begin{figure}[h!]
    \centering
    \includegraphics[scale=0.5]{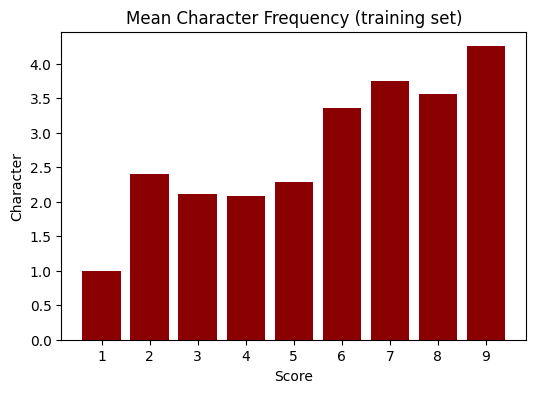}
    \includegraphics[scale=0.5]{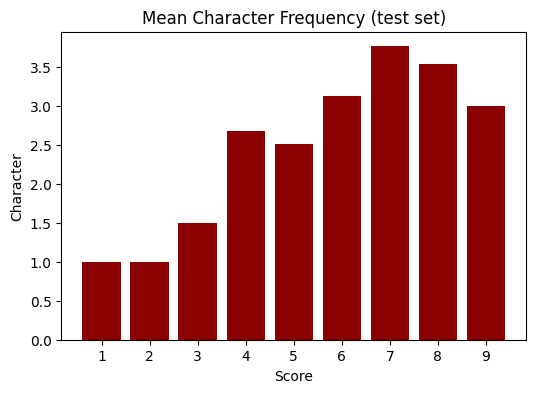}
    \caption{Mean Character Wordcount in the training (left) and test (right) set. X-axis represents the floor score. Y-axis is the mean characters per score.}
    \label{fig:traintest_char}
\end{figure}

Finally, the train-test splitting was performed randomly, but each cluster was completely contained withing a single split. The training set has $6,345$ samples ($81.5\%$) while the test set has $1,439$ samples ($18.5\%$). Fortunately, despite the random splitting, both sets obtained very similar statistics, as evidenced in Table \ref{tab:traintest_syn} and Figure \ref{fig:traintest_syn} for the synopsis wordcount, and Table \ref{tab:traintest_char} and Figure \ref{fig:traintest_char} for the character wordcount.

\begin{table}[h!]
    \centering
    \caption{Statistics for the training and test set.}
    \label{tab:traintest_syn}
    \begin{tabular}{lllll}
    \hline
    & \multicolumn{4}{c}{\textbf{Training Set}}\\
    \hline
      \textbf{Score} & \textbf{Samples} & \textbf{Max. Words} & \textbf{Min. Words} & \textbf{Avg. Words} \\
    \hline
    $1-2$   & $1\ (0.01\%)$     & $55$      & $55$      & $55$\\
    $2-3$   & $5\ (0.07\%)$     & $228$     & $25$      & $124.6$\\
    $3-4$   & $9\ (0.14\%)$     & $166$     & $24$      & $82.22$\\
    $4-5$   & $64\ (1\%)$       & $244$     & $25$      & $75.53$\\
    $5-6$   & $749\ (11.87\%)$  & $329$     & $24$      & $81.4$\\
    $6-7$   & $2,539\ (40\%)$   & $384$     & $24$      & $97.54$\\
    $7-8$   & $2,487\ (39.19\%)$& $581$     & $25$      & $121.14$\\
    $8-9$   & $483\ (7.6\%)$    & $340$     & $29$      & $151.38$\\
    $9-10$  & $8\ (0.12\%)$     & $188$     & $134$     & $157.75$\\
    \hline
    \textbf{Total}  & $6,345\ (100\%)$ & $581$ & $24$    & $105.17$\\
    \hline
    & \multicolumn{4}{c}{\textbf{Test Set}}\\
    \hline
    $1-2$   & $1\ (0.07\%)$   & $68$      & $68$      & $68$\\
    $2-3$   & $1\ (0.07\%)$   & $52$      & $52$      & $52$\\
    $3-4$   & $2\ (0.14\%)$   & $87$      & $61$      & $74$\\
    $4-5$   & $16\ (1.11\%)$  & $202$     & $27$      & $92.31$\\
    $5-6$   & $176\ (12.23\%)$& $254$     & $26$      & $92.03$\\
    $6-7$   & $592\ (41.13\%)$& $397$     & $25$      & $96.56$\\
    $7-8$   & $534\ (37.10\%)$& $277$     & $24$      & $122.07$\\
    $8-9$   & $113\ (7.88\%)$ & $234$     & $35$      & $142.39$\\
    $9-10$  & $4\ (0.27\%)$   & $189$     & $133$     & $156.25$\\
    \hline
    \textbf{Total}  & $1,439\ (100\%)$ & $397$ & $24$  & $99.51$\\
    \hline
    
    \end{tabular}
\end{table}

\begin{table}[h!]
    \centering
    \caption{Statistics for the characters.}
    \label{tab:traintest_char}
    \begin{tabular}{lllll}
    \hline
    \textbf{Split} & \textbf{Total Characters} & \textbf{Max. Words} & \textbf{Min. Words} & \textbf{Avg. Words} \\
    \hline
    Train Set & $14,168$ & $3,551$ & $4$ & $120.51$ \\
    Test Set  & $3,288$ & $2,274$ & $4$ & $124.84$ \\
    \hline
    \textbf{Total} & $14,682$ & $3,551$ & $4$ & $121.34$ \\
    \hline
    \end{tabular}
\end{table}

\subsection*{Experimental Setup}

All neural networks were implemented with PyTorch 1.10.1 \cite{paszke2019pytorch} and run on an NVIDIA Quadro RTX 6000 GPU with 46 GB of VRAM. In the case of the three-input deep neural network, the neural network used the Huggingface models \cite{huggingface-GPT2,huggingface-ResNet}. Five experiments were conducted: the benchmark with the traditional vectorizations (Trad) and the three-input deep neural network with all inputs (Full), with only synopsis (GPT-2 + MLP), only portraits (ResNET-50 + MLP), and only descriptions (GPT-2 + MLP). For synopsis and descriptions (text inputs), the large MLP depicted in Table \ref{tab:big_MLP} was modified by removing the \textit{First} layer. For portraits (image inputs), the small MLP depicted in Table \ref{tab:smol_MLP} served as the base, with two additional linear layers with ReLU activation functions and output shapes of $384$ and $1$ respectively, added for regression purposes. Finally, for the traditional methods, the MLP depicted in Table \ref{tab:trad_MLP} was used. All experiments utilized the same hyperparameters depicted in Table \ref{tab:params}, except for the epochs, since the experiment \textit{Trad} employed $30$ instead of $5$ as the other experiments. Additionally, the scores were scaled to range between $0$ and $1$, with $0$ assigned to the minimum score and $1$ to the maximum.

\subsection*{Results}

Results are depicted in Table \ref{tab:results}. Learning curves for each experiment can be found in Figure \ref{fig:lcurves}. As anticipated in the Introduction, the superior model was Full with the best metrics and the best fitting on the learning curve, without either overfitting or underfitting after the first epoch. All non-traditional experiments outperformed the MSE benchmark, but only images were not capable to surpass the benchmark correlation coefficients.

\begin{table}[h!]
    \centering
    \begin{tabular}{ll}
    \hline
         \textbf{Parameter} & \textbf{Value} \\
    \hline
         Seed & $42$. \\

         Synopsis Pretrained Model & \textit{'GPT2'}.\\
         Char. Desc. Pretrained Model & \textit{'GPT2'}.\\
         Image Pretrained Model & \textit{'microsoft/resnet-50'}.\\
         Synopsis Tokenizer Max. Length & $128$ (\textit{GPT2tokenizer}).\\
         Char. Desc. Tokenizer Max. Length & $256$ (\textit{GPT2tokenizer}).\\
         Image Processors Parameters & Default (\textit{AutoImageProcessor}).\\
    \hline
         
         Optimizer & Adam with weight decay (AdamW).\\
         Opt. Learning Rate & $5e-2$.\\
         Opt. Epsilon param & $1e-8$.\\
    \hline
         Loss Function & Mean Squared Error (MSEloss). \\
         Batch Size & $16$. \\
         Epochs & $5$ ($30$ for traditional vectorizations).\\
    \hline
    \end{tabular}
    \caption{All employed parameters in the experiments.}
    \label{tab:params}
\end{table}

\begin{table}[h!]
    \centering
    \begin{tabular}{lllllll}
    
    \hline
    \multicolumn{6}{c}{\textbf{Popularity Prediction}}\\
    \hline
        \textbf{Architecture} & \textbf{Input} & \textbf{MSE} & \textbf{Spearman} & \textbf{Pearson} &\textbf{Kendall's Tau} \\
    \hline
        Full & All & $0.011$ & \textbf{0.431} & \textbf{0.436} & $0.297$ \\
        GPT-2+MLP & Syn. & $0.012$ & \textbf{0.338} & \textbf{0.328} & $0.230$ \\
        GPT-2+MLP & Char. (Desc.) & $0.012$ & \textbf{0.307} & \textbf{0.341} & $0.210$ \\
        ResNET-50+MLP & Char. (Img.) & $0.028$ & $0.096$ & $0.121$ & $0.065$ \\
        \hline
        Trad (benchmark) & All & $0.412$ & $0.195$ & $0.183$  & $0.130$\\
    \hline
    
    \end{tabular}
    \caption{All experiments, sorted by mean squared error (MSE): the lower the value, the best the result. Moderate correlations are highligted with \textbf{bold}.}
    \label{tab:results}
\end{table}

\section*{Discussion}\label{Sec:discussion}

Recall that the MSE is calculated with the square values of the differences between the real and the predicted scores. Hence, a value closer to $0$ might seem as the \textit{best} scenario possible, as long as the learning curve does not evidence overfitting or underfitting.

\begin{figure}[h!t!]
    \centering
    \includegraphics[scale=0.47]{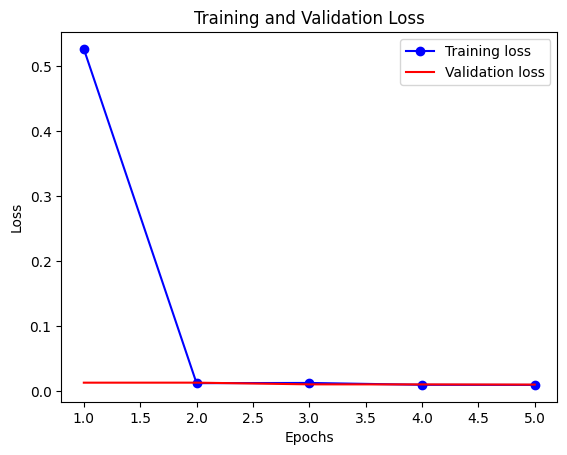}
    \includegraphics[scale=0.47]{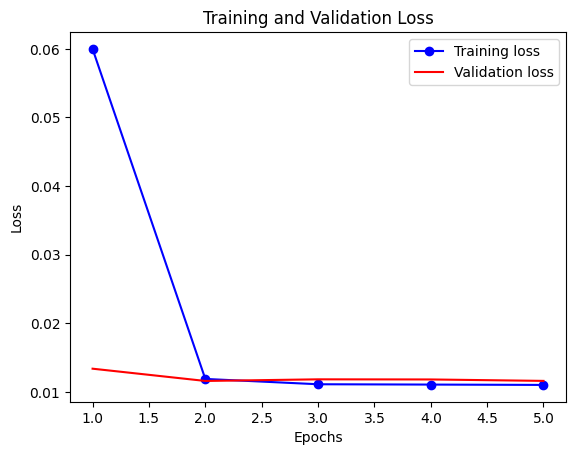}
    \includegraphics[scale=0.47]{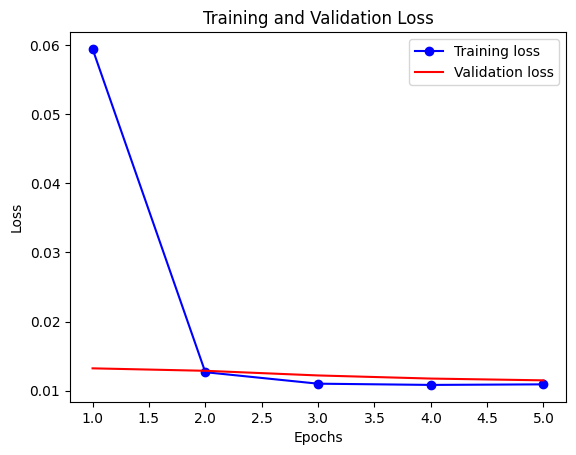}
    \includegraphics[scale=0.47]{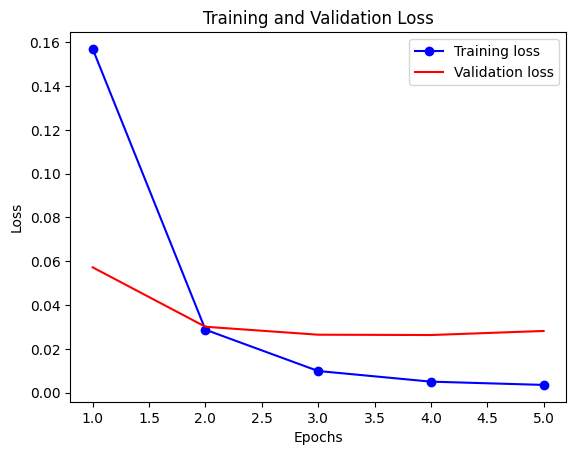}
    \includegraphics[scale=0.47]{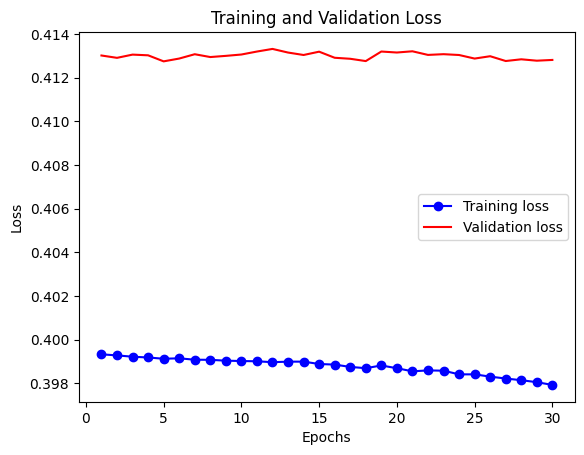}
    \caption{Learning curves for all experiments. From top-left to bottom-right: All inputs (good fit), Syn+MLP (slightly underfit), Char+MLP (slightly underfit), Img+MLP (notable underfit) and traditional methods (catastrophic underfit, as expected due the size and complexity of the dataset).}
    \label{fig:lcurves}
\end{figure}

As evidenced in Figure \ref{fig:lcurves}, all models with not full inputs evidenced underfitting, being the images (the smallest model) the most critical case. Recall that underfits happens when the model is not big enough to capture the complexity of the dataset. Hence, this learning curves confirms the hypothesis stated in the introduction about the requirement of larger, deeper and complex models. This is also promising about the employment of LLMs such as Llama2 \cite{Touvron2023Llama2O} or GPT4 \cite{achiam2023gpt}.

\subsubsection*{Correlations}

According to \cite{cohen1992statistical}, it is possible to interpretate the correlation values according to the follow rule of thumb: $0-0.29$ indicates a small correlation, $0.30-0.49$ indicates a moderate correlation and $0.50-1$ indicates a strong correlation. Fine-graining, in \cite{Abdalla2021WhatMS}, the authors used another rule of thumb for the Spearman correlation coefficient: $0–0.19$: very weak; $0.2–.39$: weak; $0.4–0.59$: moderate; $0.6–0.79$: strong; $0.8–1$: very strong. In this work, the follow \textit{hybrid} rule of thumb is considered:
\begin{itemize}
    \item $0-0.19$ for a very weak correlation.
    \item $0.2-0.29$ for a weak correlation.
    \item $0.3-0.49$ for a moderate correlation.
    \item $0.5-1$ for a strong correlation.
\end{itemize}

All the inputs together demonstrated a moderate correlation with the MAL score. Images showed the weakest correlation, supporting more the need of larger models. Not surprisingly, only text-based inputs showed a similar moderate correlation.

Fine-grained analysis of each metric shows that the moderate Pearson and Spearman correlations with text-based inputs suggests that the latent space generated by the larger MLP was moderately successful in linearizing the problem, indicating that it may be possible to completely linearize it with more complex and rich models. Adding images significatively strengthens this linear relationship (adding up $0.1$ points), showing that main character visual information may be crucial for popularity. However, it is noteworthy that images by themselves had a very weak correlation, underscoring the essential role of text descriptions. Spearman, with a similar behaviour, also adds ranking information.

Finally, the Kendall's Tau correlation helps to better understand the dataset as a whole. it is noteworthy that the minimum score on the dataset is $1.86$ while the maximum is $9.06$. The significantly lower results concerning this metric suggest that these outliers have a considerable impact during the training process. In Figure \ref{fig:stats_full}, the statistical difference between these outliers is evident, as well as in Table \ref{tab:stats_syn}, where the two samples with scores between $1$ and $2$ have $55-68$ words, while the twelve samples with scores between $9$ and $10$ have $133-189$ words. These findings encourage the use of class weights in the logits to address such score outliers and alleviate related problems. However, this measure might be artificial and can lead to unexpected biases, as it is a product of a weakness in the dataset, so it was not implemented in this work.

\subsubsection*{A Memory Constraint with Transformer-based Models}

Although the correlation metrics showed that main character descriptions have a considerable correlation with the labels, the proposed GPT-2 model presents the second most pronounced underfitting, only behind the main character portraits. Table \ref{tab:params} shows that the maximum length for the tokenizer of main character descriptions is $256$. However, some of these descriptions, when concatenated, can surpass $1,000$ words, as shown in the example in Table \ref{fig:merged}. Consequently, this underfitting can be explained by the loss of crucial information.

Due to the memory limitations of transformer-based models, it is not trivial to increase the tokenizer size. Therefore, further work should explore more memory-efficient methods capable of capturing all information with similar learning capabilities (e.g., Mamba \cite{gu2023mamba}) or find other alternatives to include all main character information simultaneously (e.g., generating mean tensors and assessing how much relevant information is truly lost from doing that).

\section*{Conclusions}\label{Sec:concl}

In this paper, one of the most robust free datasets for anime popularity prediction is introduced, relying solely on freely available internet data. To explore this dataset and propose a model for solving the task, a deep neural network leveraging GPT-2 and ResNET-50 was developed. The best MSE achieved was $0.011$, which is significantly lower than the benchmark MSE of $0.412$.

Moderate correlations were achieved with Spearman and Pearson metrics, while a low correlation with Kendall's Tau revealed important quirks in the dataset, such as marginal representativity of animes with extreme scores. These issues should be taken into account when designing future models to solve this task, and also aligns with the claims in \cite{Yu2023SelfSupervised} that recommender systems can be significantly enhanced by solving the sparsity problem in data, also presents in non-free datasets.

The correlation coefficients also demonstrated that adding character portraits significantly enhances text-based inputs, even when embedded with a relatively small model such as ResNET-50.

Further work can be summarized as follows:
\begin{itemize}
    \item \textbf{The memory problem with Main Character Descriptions}: Truncating main character descriptions to $256$ tokens significantly impacts the results due the information lost, reflected in the underfitting. This problem, closely related with the memory constraint of transformer-based models, have these two paths to overcome it:
    \begin{itemize}
        \item To experiment with other language models with less memory requirements. E.g. n-grams vectorizations or selective state spaces (e.g. Mamba).
        \item To adjust the input to embed each main character individually. These individual tensors can then be processed in a miriad of ways, such as averaging them. The information lost after such processes is an important topic to be explored.
    \end{itemize}

    \item \textbf{Larger models for processing the inputs}: GPT-2 can be upgraded to GPT-4 or Llama2, and ResNET-50 can be enhanced to its bigger brother ResNET-152, a very deep pretrained CNN or a vision transformer such as Image-GPT. The moderate Pearson and Spearman coeficients indicates that leveraging larger models will significantly benefit the results.
    
\end{itemize}

\section*{Acknowledgments}

The work was done with partial support from the Mexican Government through the grant A1-S-47854 of CONACYT, Mexico, grants 20241816, 20241819, and 20240951 of the Secretaría de Investigación y Posgrado of the Instituto Politécnico Nacional, Mexico. The authors thank the CONACYT for the computing resources brought to them through the Plataforma de Aprendizaje Profundo para Tecnologías del Lenguaje of the Laboratorio de Supercómputo of the INAOE, Mexico and acknowledge the support of Microsoft through the Microsoft Latin America PhD Award.

\bibliographystyle{unsrt} 
\bibliography{templateArxiv.bib}

\end{document}